\documentclass[a4paper,fleqn]{cas-dc}

\usepackage[authoryear]{natbib}
\usepackage{graphicx}
\usepackage{comment}
\usepackage{amsmath,amssymb} 
\usepackage{color}
\usepackage{epsfig}
\usepackage{graphicx}
\usepackage{algorithm}
\usepackage{algpseudocode}
\usepackage[percent]{overpic}
\usepackage{boldline}
\usepackage{multirow}
\usepackage{calligra}
\usepackage{url}

\usepackage{epstopdf}
\usepackage{makecell}

\usepackage{lipsum}
\usepackage{subfig}
\usepackage{tabu}
\usepackage{mathtools}

\usepackage{booktabs,arydshln}

\def\tsc#1{\csdef{#1}{\textsc{\lowercase{#1}}\xspace}}
\tsc{WGM}
\tsc{QE}
\tsc{EP}
\tsc{PMS}
\tsc{BEC}
\tsc{DE}

\begin{document}
\let\WriteBookmarks\relax
\def\floatpagepagefraction{1}
\def\textpagefraction{.001}
\shorttitle{Auto-Augmentation for n-Shot Learning}
\shortauthors{A. Naghizadeh et~al.}

\title [mode = title]{Greedy AutoAugment}

\author[1]{Alireza Naghizadeh} 
\cortext[cor1]{Corresponding author:}
\ead{ar.naghizadeh@cs.rutgers.edu}
\author[2]{Mohammadsajad~Abavisani}
\author[1]{Dimitris~N.~Metaxas}

\address[1]{Department of  Computer Science, Rutgers University, CBIM, Piscataway Township, NJ 08854.}
\address[2]{Department of Electrical and Computer Engineering, Georgia Institute of Technology, TReNDs center, 55 Park Pl NE, Atlanta, GA 30303.}

\cortext[cor1]{Corresponding author}

\begin{abstract}
A major problem in data augmentation is to ensure that the generated new samples cover the search space. This is a challenging problem and requires exploration for data augmentation policies to ensure their effectiveness in covering the search space. In this paper, we propose  Greedy AutoAugment as a highly efficient search algorithm to find the best augmentation policies. We use a greedy approach to reduce the exponential growth of the number of possible trials to linear growth. The Greedy Search also helps us to lead the search towards the sub-policies with better results, which eventually helps to increase the accuracy. The proposed method can be used as a reliable addition to the current artifitial neural networks. Our experiments on four datasets (Tiny ImageNet, CIFAR-10, CIFAR-100, and SVHN) show that Greedy AutoAugment provides better accuracy, while using 360 times fewer computational resources.\end{abstract}


\maketitle
\section{Introduction}

Data augmentation \footnote{This is the the arxiv version of the Greedy AutoAugment paper \citep{naghizadeh2020greedy}.} is an important technique that can help to improve the performance of various data analysis algorithms in the presence of insufficient data. For instance, a common practice in medical applications is class-specific data augmentation. In this case, gathering sufficient labeled data to train a  deep neural network model is impractical  \citep{ronneberger2015u, kuo2019automation}. This is a classic type of long-tail distribution data, which is also prevalent in natural images \citep{zhu2014capturing, tang2015face} and can be addressed by data augmentation methods \citep{wang2017learning}. Other important usages of data augmentation methods include unsupervised learning  \citep{NIPS2014_5548,devries2017dataset,bengio2013better, ozair2014deep,naghizadeh2020gnm,naghizadeh2020condensed} and in the improved training of generative adversarial networks \citep{peng2018jointly,luan2017deep}. In this paper, we focus on the problem of data augmentation for image classification .  The main goal is to increase the accuracy of image classification by applying the right augmentation techniques on training data.

Data augmentation in image classification is directly related to image-based object transformations. In the classification process,  it is desirable to take into account a variety of such transformations to improve image classification. In other words, we want the perception of an object to be invariant to the properties such as scale, brightness, rotation, and viewing angle.  Estimating important object transformations and applying them in the learning process is a critical problem in Artificial Neural Networks (ANNs). For instance,  it is desirable that a network, after learning an object from its original form, recognizes the same object in a modified location where its scale, rotation, and other properties have been changed.  Currently, there are two ways to deal with this problem. First, by designing network architectures that can inherently be invariant to important image-based object transformations. Second, with the use of data augmentation methods.

The most basic network which considers the transformations of the input data is the Convolutional Neural Network (CNN). The CNN architecture, with the concept of convolutional layers, tries to be translation invariant  \citep{lecun1990handwritten, krizhevsky2012imagenet}. This network was very successful in its approach and has been used as a basis for the development of more advanced architectures  \citep{szegedy2015going,he2016deep,huang2017densely}. Another example of this approach is CapsuleNet, which tries to find the relevant pose information automatically  \citep{hinton2011transforming,hinton2018matrix,sabour2017dynamic}. While the design of CapsuleNet improved the results of basic datasets  \citep{lecun1998mnist}, unfortunately, it could not improve the accuracy for more complicated datasets such as ImageNet \citep{imagenet_cvpr09}.

The second method for considering different transformations of the input data is to use data augmentation. In this method, the objective is to achieve invariance by applying different image transformations such as geometry transformation, kernel filtering, color transformation, image mixing, random erasing \citep{devries2017improved}, etc. The main advantage of this method is simplicity and supporting all forms of ANN architectures.  Additionally, there is a possibility to use transformation techniques in which current ANN architectures do not support. 

One of the most important factors for data augmentation techniques is the constraint on the number of possibilities for applying augmentation techniques. In this regard, only a subset of the possible techniques can be used for data augmentation.  Therefore,  a search mechanism is needed to find the best possible techniques. The most common method to find the best data augmentation techniques is to find them manually \citep{krizhevsky2012imagenet, cirecsan2012multi, huang2016deep} which needs prior knowledge and expertise. Recently,  the AutoAugment \citep{cubuk2018autoaugment} is proposed to automate the process of finding the best augmentations.  In this method, finding the augmentation policy is reduced to a discrete search problem over various augmentation techniques, each having hyperparameters of the probability of applying the operation and the magnitude to which the operation is applied. Because of the computational requirements for searching with AutoAugment, this method relies on transferability of the augmentation techniques.  This means that the policies that are found with one dataset and architecture can be used for similar datasets with different architectures \citep{cubuk2018autoaugment}. 

Using the same policies which are found for a specific scenario and trying to use them for others is risky and may lead to worse results (see the Results). To solve this problem, it is desired to perform the search more effectively, so that searching can be performed for each dataset separately. In this paper, we address the problem of how to effectively search for the most appropriate data augmentation techniques and apply them to training data. For this purpose, we propose Greedy AutoAugment. In this method, we develop a greedy-based search algorithm, which reduces the searching space from the exponential growth of possible trials to linear growth. With the proposed method, instead of searching for all possible outcomes, the search expands dynamically, and for each sub-policy, it is pushed towards trials with the best outcomes. To achieve this goal, we search among sub-policies with only two elements, the techniques of operations and the magnitudes for those techniques. The probability is applied after the search is finished with a methodology that gives weights to better policies. Our experimental results show that this approach is effective in providing higher accuracies.  The Greedy AutoAugment, on four datasets,  Tiny ImageNet, CIFAR-10, CIFAR-100, and SVHN, could reliably provide higher accuracies while using 360 times fewer computational resources.

\section{Data Augmentation}

Data augmentation refers to the practice of applying a series of standard transformations \citep{wiredfool_2016_44297, devries2017improved, inoue2018data} to the given image data. In order to use these transformations, for each epoch in the training phase, a percentage of the data receives one or a combination of these techniques.

\begin{figure}[h]
	\centering
	\subfloat[ImageNet] {
		\includegraphics [width = 28mm]{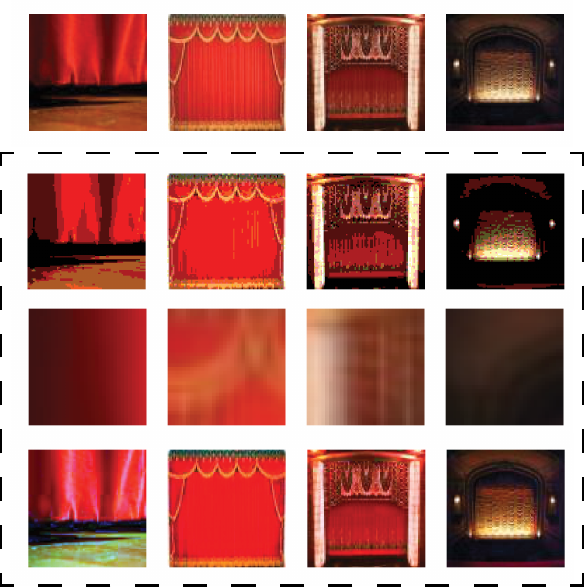}
	}\hspace{5mm}
	\subfloat[CIFAR-100]{
		\includegraphics [width = 28mm]{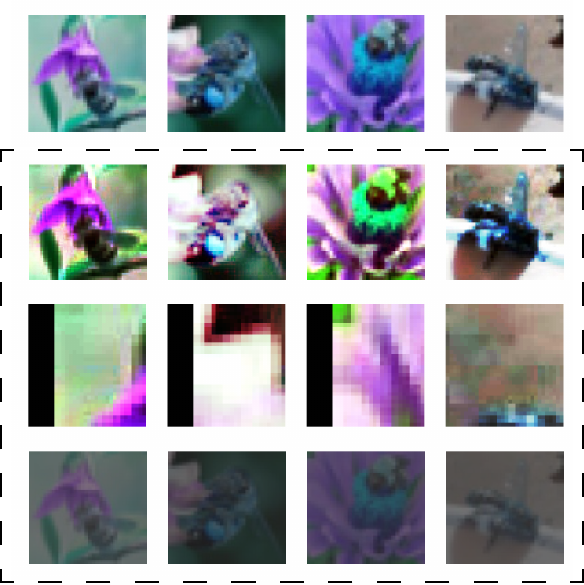}
	}
	\caption{Random data augmentation techniques applied to samples from two real datasets. Each row receives the same augmentation technique with different magnitudes.}
	\label{fig0}
\end{figure}

The effect of applying data augmentations to the images from ImageNet and CIFAR-100 datasets are shown in Figure \ref{fig0}. In each row, one specific combination of augmentation techniques with different magnitudes is chosen randomly and is applied to the columns of images. As we can see, in some instances, the change is not noticeable, and in others, the change could completely change the original data. The goal is to discard combinations that would decrease the generalization ability of the network and select the best combinations that would increase it. 

The augmentation techniques in their simplest form have been used in prominent artificial neural networks \citep{krizhevsky2012imagenet, cirecsan2012multi, huang2016deep}.  These networks mostly used a  trial and error approach to manually find the best combinations. A more advanced approach is proposed in \citep{cubuk2018autoaugment}, which automates the searching process for finding the best augmentation techniques.  There are also methods that do not perform a real search to find the best augmentations. Instead, these methods are created to be resilient against randomly selected augmentations \citep{ho2019pba, lim2019fast}.  

\begin{figure}[h]
	
	\includegraphics [width = 75mm]{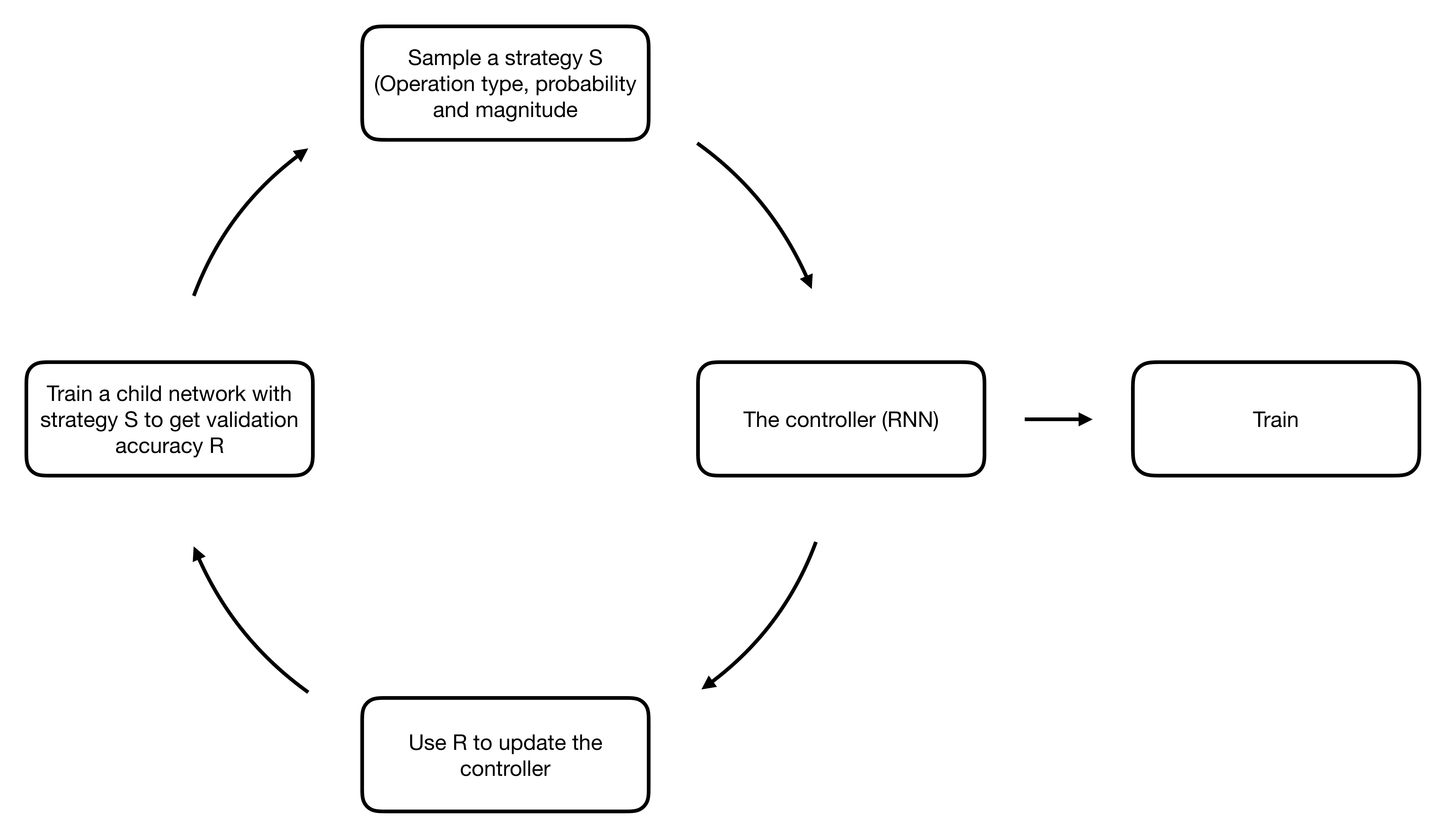}
	
	\caption{The general scheme of AutoAugment algorithm. }
	\label{figvv0}
	
\end{figure} 

\begin{table*}[h]
\centering
\caption{Augmentation techniques with their descriptions which are used in GAutoAugment.}
\resizebox{140mm}{!}{
\begin{tabular}{@{}cll|cll@{}}

\cmidrule(l){1-6}
\quad & Technique  &  Description &  & Technique  &  Description    \\
 \cmidrule(l){1-6}

\multirow{4}{*}
 \quad 1. & FlipLR	& Filliping the image	along the vertical axis. & 11. & Contrast&	Changing the contrast of the image.\\
 \cmidrule(lr){1-6}
 \quad 2. & FlipUD & 	Filliping the image	along the horizontal axis.	&  12. & Brightness& Adjusting the brightness of the image. \\
 \cmidrule(lr){1-6}
 \quad  3.& AutoContrast	& Increasing the contrast of the image. & 	 13. & Sharpness& Adjusting the sharpness of the image.\\
 \cmidrule(lr){1-6}
 \quad  4. & Equalize	& Equalizing the histogram of the image. &  14. &	ShearX& Sheering the image in horizontal axis. \\
 \cmidrule(lr){1-6}
 \quad  5. & Invert	& Inverting the color of the pixels in the image. 	&  15. & ShearY & Sheering the image in vertical axis. 	\\
\cmidrule(lr){1-6}
\quad  6. & Rotate	& Rotating the image by certain degrees.	&  16.& TranslateX&	Translating the image in horizontal axis.\\
 \cmidrule(lr){1-6}
 \quad  7. & Posterize & 	Redicing the number of Bits for each pixel.	&  17.& TranslateY& Translating the image in vertical axis.\\
 \cmidrule(lr){1-6}
 \quad  8. & CropBilinear	& Croping with bilinear interpolation strategy.& 	 18.& Cutout & Changing a random square patch of the image to gray pixels.\\
 \cmidrule(lr){1-6}
 \quad  9. & Solarize	& Inverting the color of all the pixels above a certain threshold. &  19.& Blur& Blurring the image. \\
 \cmidrule(lr){1-6}
 \quad  10. & Color	& Changing the color balance of the image.	&  20.& Smooth & Smoothing the image(Low-pass filtering).	\\
\cmidrule(lr){1-6}
 
\end{tabular}
}
\label{tbl3}
\end{table*}

The searching process of the AutoAugment method heavily relies on NasNet  \citep{zoph2018learning} as a controller to direct the search. The controller predicts a decision by using a one-layer LSTM, which contains 100 hidden units and $30$ units softmax predictions. The prediction is then fed into the next step as an embedding. In the end, the controller uses 30 softmax predictions in order to select the best policies. The LSTM should be in direct contact with some child networks to train its own network.  The child networks are usually a subset of the training networks. The accuracy updates from child networks are used to update the main LSTM network. To use the accuracies, a policy gradient method called Proximal Policy Optimization algorithm (PPO) is employed. The whole process is shown in Figure \ref{figvv0}. The strategies (policies) have three elements 1- data augmentation techniques, 2- the probability of applying each operation, and 3- the magnitude of the operation. The policy gradient receives the values of reward R to update the controller. The LSTM uses the softmax predictions of its trained networks to select the best polices.  In the end, the best policies are used to train the actual network.

\section{Searching Environment}

To perform data augmentation on image data, we use policies. Let us define policy function $\hat {f}_i$ that may include one or more sub-policy functions. The sub-policy is used as the functionality of applying augmentation techniques on image data. Each sub-policy has three essential elements, 1- the augmentation technique, 2- the magnitude of the operation, and 3- the probability of applying the operation. For a complete list of augmentation techniques that we use in this paper, see Table \ref{tbl3}. The magnitude is the degree in which an operation is applied. For instance, in the rotate augmentation, the magnitude specifies how much we should rotate an image. The third element specifies the probability of applying the augmentation on the image.   

We define each image as a multivariate data point $\textbf{x}$. For augmentation on image data we use $\hat{f}(\textbf{x})\rightarrow \textbf{x}'$ which transforms the original image $\textbf{x}$ into the augmented image $\textbf{x}'$. The $\hat{f}_i$ is separated into one or a multiple number of sub-policies. The sub-policy function is denoted by $f$. Each sub-policy receives three input values, $f(t,p,m)$, where the $t$, $p$, and $m$ variables represent the augmentation technique, probability, and magnitude. The search space is defined as all of the possible combinations of concatenated sub-policies. The search space for a single $f$, includes all of the possible combinations of discrete values of $t,p$ and $m$. Accordingly, the search space for any $f$ is $(t_n \times p_n \times m_n)$ where $t_n,p_n$, and $m_n$ are the maximum values for $t,p$, and $m$.

To represent the search space, we can divide it into different layers. A search space that has only one layer includes an augmentation function $\hat{f}_i$ that has only one $f$ as its sub-policy. We know that the search space for sub-policy $f$ has a maximum number of possible elements of $(t_n \times p_n \times m_n)$ . The search space with two layers is defined for $\hat{f}_i$s that can concatenate two $f$s as their sub-policies. Respectively, the maximum number of possible elements for the search space expands to $(t_n \times p_n \times m_n)^2$. We generalize the number of layers for any search space with the variable $\ell$. The $\ell$ is a discreet integer value such that $\ell \geq 1$. The search space is defined as concatenated layers, with a maximum size of $(t_n \times p_n \times m_n)^\ell$, where $\ell$ indicates the number of allowed sub-policies for $\hat{f}_i$.

The important point in the defined searching environment is that the search space expands exponentially. Particularly,  given base $ (t_n \times p_n \times m_n)$ the possible number of trials increases exponentially with the increase of $\ell$. To solve this problem, we transform the base of the search space into a two-variable setup. Among the three variables, $t, p$, and $m$ in the sub-policies, we do not search for the variable $p$ and fixate its value to one. Instead of searching for the best probability values among sub-policies, we apply the probability with a methodology that gives more weights to the policies with better accuracy results. This simple design decision helps us to reduce the computational requirements of the searching algorithm considerably. 

Even with reducing the number of variables for the base, the growth of the number of possible trials is still exponential. To tackle this problem, when going from one layer to the other (concatenating two sub-policies), we use a greedy search. To use the greedy search, we replace the reinforcement learning in AutoAugment with  Breadth-First Search, which is an explorer for tree-based data structures. Accordingly, the Greedy Breadth-First Search is used to explore the defined layered environment. The new number of the possibilities is defined as follows, 

\begin{align}\label{eq:DSC}
\sum _{ 1 }^{ k }{ (t_n \times m_n) } 
\end{align}

In this notation, $k$ is an arbitrary integer number, which indicates the number of iterations the algorithm is allowed to perform in the search.  The higher values of the $k$, allow algorithm to perform more trials, which may lead to higher accuracy. Therefore, the value of $k$ should be set based on the available computational resources.  The greedy approach helps us to convert the exponential growth of base $(t_n \times p_n \times m_n)$ with $\ell$ to linear growth of base $(t_n \times m_n)$ times the value of $k$ which allows $\hat{f}_i$ to potentially encompass many layers. For numerical values, we follow the discretization setup, which is introduced in AutoAugment.  The number of augmentation techniques that we use in this paper is $t_n=20$. The discretization of probabilities is with $p_n=11$ values with uniform space, and the discretization of magnitudes is with $m_n=10$ values with uniform space.

\section{Greedy AutoAugment}

Training with the Greedy AutoAugment algorithm includes two main steps. The first step is to search and find the best $\hat{f}_i$s. The second step is to apply the $\hat{f}_i$s to $\textbf{X}$ and increase the generalization ability of the network. To perform the greedy search, we use Algorithm \ref{alg1}.  The input of the searching process is $\textbf{X}$, and the output is $(\hat{f}_1,\dots,\hat{f}_d)$, which are the best augmentation policies found in the search. The number of iterations of the algorithm needs to be at least equal to one ($k \ge 1$).

The lines 1-10 are for $k=1$. The Breadth-First Search takes place in lines 1,2, where we go through all of the augmentation techniques and their respective magnitudes.  In line 3, we use fixed value of $1$ for the probability. In line 4, $\textbf{X}$ is divided into two parts $\textbf{X}_{tr}, \textbf{X}_{te}$ for training and testing. In line 5, the sub-policy is considered as a complete policy. In line 6, sub-policy f(t,p,m) are applied on $\textbf{X}_{tr}$. The classification of $\textbf{X}_{te}$ is determined in line 7 and the score of the sub-policy and its respective score is stored.  Line 9 is for the greedy part of the algorithm, where the policy with the highest score is stored to act as a base sub-policy for more iterations.

\begin{algorithm}[t!]
	\caption{The proposed Greedy AutoAugment algorithm.}
	\label{alg1}
	\begin{algorithmic}
		\State \textbf{Input:} Dataset $\textbf{X}$.
		\State \textbf{Output:} Best augmentation policy functions \textbf {$\hat{f}_1,\dots,\hat{f}_d$}.
		\begin{algorithmic}[1]
			\For {all augmentation techniques $t$ = 1,\dots, $t_n$}
			\For {all magnitudes $m$ = 1, \dots , $m_n$}
			\State{set probability to 1;}
			\State{$\textbf{X} \rightarrow \textbf{X}_{tr} ,  \textbf{X}_{te}$;}
			\State{$\hat{f} = f(t,p,m)$;}
			\State{apply $\hat{f}$ on $\textbf{X}_{tr}$;}
			\State{store policy with its score on $\textbf{X}_{te}$.}
			\EndFor
			\State{store best policy}
			\EndFor
			\For {counter = 2,\dots, $k$}
			\State{$\hat{f}$ = best policy;}
			\For {all augmentation techniques $t$ = 1,\dots, $t_n$}
			\For {all magnitudes $m$ = 1, \dots , $m_n$}
			\State{set probability to 1;}
			\State{$\textbf{X} \rightarrow \textbf{X}_{tr} ,  \textbf{X}_{te}$;}
			\State{$\hat{f} = \hat{f} + f(t,p,m)$;}
			\State{apply $\hat{f}$ on $\textbf{X}_{tr}$;}
			\State{store policy with its score on $\textbf{X}_{te}$.}
			\EndFor
			\EndFor
			\State{store best policy}
			\EndFor
			\State{select $\hat{f}_1,\dots,\hat{f}_d$ from all policies which have highest scores;}
			\State{ \textbf {return $\hat{f}_1,\dots,\hat{f}_d$}}
		\end{algorithmic}
	\end{algorithmic}
\end{algorithm}		

For the next step (lines 11-23) the algorithm is for iterations with $k > 1$. As it is shown in line 11, the algorithm repeats its steps until the counter is equal to $k$. In line 12, the $\hat{f}$ is equal to the best policy with the highest score, which is not selected before. In this way, the next iteration starts from a base that has the best score. As we can see in line 17, the f(t,p,m) is added on top of the best policy. The other parts of the algorithm are the same as the first part. In the end, we select the $d$ best $\hat{f}_i$s, which shows the highest accuracy results among all policies. In Algorithm \ref{alg1}, the probabilities for all policies were always set to the value of one. To determine the $p$ values for $\hat{f}_i$s we propose the following process. 

The goal is that the policy which has a higher score to receive a higher probability. For this purpose, let us define $S_1, \dots, S_{d/\lambda}$, where $S_i$, is a set with $\lambda$ number of probabilities, $S_1=\{p_1,\dots,p_\lambda\}, \dots ,S_{d/\lambda}=\{p_{d-\lambda},\dots,p_d\}$. In this regard,  we define  vector $\vec { v } =[v_1, \dots , v_{d/\lambda},v_{{(d/\lambda)}+1}]$ which represents the probability of choosing a set $S_i$ or the original data. The $v_1$ corresponds to selecting the original data and $v_2,\dots, v_{d/\lambda},v_{{(d/\lambda)}+1}$ correspond to the selection of $S_1, \dots, S_{d/\lambda}$. To fill the values of $\vec { v }$, we use the Pareto Distribution \citep{arnold2015}, as follows,
	
	\begin{equation}
	\label{eqn4-1}
	\rightarrow v_i =
	\left\{
	\begin{array}{l l}
	(\frac{1}{i}) ^ \alpha \quad\quad\quad    i > 1\\
	\quad1  \quad\quad\quad\   i \le 1
	\end{array}\right.
	\end{equation}
	
In this equation,  a positive parameter ($\alpha$)  is used to assign the highest probability to $v_1$ and lowest probability to $v_{{(d/\lambda)}+1}$.  When $i=1$ the probability is one, and for $i > 1$ the probability is less than one but not zero. To choose the best set $S_i$ or original data, we start from the rightmost element of $\vec{v}$ and go to the leftmost element of $\vec{v}$. Each of these elements has a chance to be selected based on their respective $v_i$ values. After choosing the $S_i$, a policy is chosen randomly from the members of the set, with uniform distribution.

\begin{figure*}
\centering
 \subfloat[]{
 \label{fig3a}
     \includegraphics [width = 28mm]{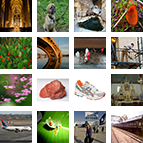}
  }\hspace{3mm}
 \subfloat[]{
 \label{fig3b}
     \includegraphics [width = 28mm]{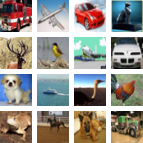}
  }\hspace{3mm}
 \subfloat[]{
 \label{fig3c}
     \includegraphics [width = 28mm]{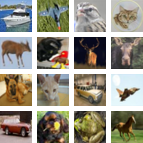}
  }\hspace{3mm}
   \subfloat[]{
 \label{fig3d}
     \includegraphics [width = 28mm]{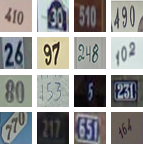}
  }

\caption{Samples from real datasets used in our experiments: (a) Tiny ImageNet (b) CIFAR-10 (c) CIFAR-100 (d) SVHN.}
\label{fig3}

\end{figure*}

\section{Results}

In this section, we compare our method with current solutions. For this purpose, first, we show the accuracy results of our method compared to the other methods. Next,  we provide a computational analysis of the overall augmentation process. The AutoAugment paper uses state-of-the-art record-breaking networks also to test their method. This encompasses several architectural advancements, which increases RAM requirements in GPUs and require at least 1800 epochs for each test-case. Using such a vast requirement to test augmentation policies is actually not necessary and would limit us to analyze the methods thoroughly. Therefore, in order for the experiments to be in our available resources, we have used a smaller and reasonable infrastructure to test both our methods and the AutoAugment results in a uniform setting. According to the AutoAugment paper, their method is transferable. Therefore, this change is fair and should not affect their method.

In these experiments, we use eleven different prominent ANN architectures.  The DenseNet121 \citep{huang2017densely}, GoogLeNet \citep{szegedy2015going}, MobileNet \citep{howard2017mobilenets}, MobileNetV2 \citep{sandler2018mobilenetv2}, PreActResNet18 \citep{he2016identity}, ResNet18 \citep{he2016deep},  ResNeXt29 \citep{xie2017aggregated}, ShuffleNetG2 \citep{zhang2018shufflenet}, ShuffleNetV2 \citep{ma2018shufflenet},  and VGG \citep{simonyan2014very} are used as prominent networks which contain both normal and lightweight networks.  To implement these networks, we forked the implementations from  \citep{kuangliu}. The default settings of the network implementations are not changed. The networks accept $32 \times 32$ images and provide an output based on the number of classes. 

\newlength{\longestentry}
\setlength{\longestentry}{\widthof{Network}}

\newlength{\mytablewidth}
\setlength{\mytablewidth}{\textwidth-\longestentry-2\tabcolsep-2\arrayrulewidth}

\newcolumntype{T}{>{\centering\arraybackslash}p{0.33\mytablewidth-2\tabcolsep-\arrayrulewidth}}
\newcolumntype{F}{>{\centering\arraybackslash}p{0.25\mytablewidth-2\tabcolsep-\arrayrulewidth}}

\begin{table*}[h]
	\centering
	\caption{The result table for accuracy analysis. Abbreviations include: Manual = Manual Augmentation, GAutoAugment = Greedy AutoAugment. }
	
	\resizebox{130mm}{!}{
		\noindent
   \begin{tabular}{l|cccc|cccc|cccc}
      \hline
     \multicolumn{2}{l}{ \multirow{2}{*}{Network} }&
      \multicolumn{4}{F}{Manual} &
      \multicolumn{4}{F}{AutoAugment} & 
      \multicolumn{3}{F}{GAutoAugment}\\
      \cline{2-13}
 &  \multicolumn{1}{c}{Tiny} &\multicolumn{1}{c}{Cifar10} &\multicolumn{1}{c}{Cifar100} &\multicolumn{1}{c}{SVHN} &\multicolumn{1}{c}{Tiny} &\multicolumn{1}{c}{Cifar10} &\multicolumn{1}{c}{Cifar100} &\multicolumn{1}{c}{SVHN} &\multicolumn{1}{c}{Tiny} &\multicolumn{1}{c}{Cifar10} &\multicolumn{1}{c}{Cifar10} &\multicolumn{1}{c}{SVHN} \\
\cline{1-13} 
      DenseNet121 	&	34.65	&	80.67	&	53.91	&	91.69	&	29.15	&	80.20	&	54.80	&	91.47	&	\textbf{37.75}	&	\textbf{83.36}	&	\textbf{59.32}	&	\textbf{93.26}\\
      GoogLeNet &	\textbf{33.78}	&	79.85	&	38.96	&	89.63	&	27.81	&	\textbf{79.92}	&	\textbf{53.28}	&	90.79	&	28.70	&	79.25	&	50.51	&\textbf{91.44}	\\
      MobileNet 	&	14.58	&	\textbf{68.50}	&	40.75	&	81.43	&	10.50	&	63.95	&	37.28	&	79.53	&	\textbf{17.38}	&	64.90	&	\textbf{43.09}	&	\textbf{88.55}	\\
      MobileNetV2 	&	\textbf{21.66}	&	70.81	&	39.92	&	84.33	&	15.00	&	67.95	&	39.14	&	83.42	&	21.07	&	\textbf{73.67}	&	\textbf{44.92}	&	\textbf{88.17}	\\
      PreActResNet18 	&	29.45	&	77.16	&	44.57	&	89.06	&	26.01	&	83.05	&	\textbf{53.28}	&	89.77	&	\textbf{32.60}	&	\textbf{83.43}	&	48.26	&	\textbf{93.60}	\\
      ResNet18 	&	29.61	&	78.68	&	42.34	&	87.08	&	27.51	&	80.22	&	\textbf{55.29}	&	89.83	&	\textbf{35.84}	&	\textbf{80.62}	&	48.25	&	\textbf{93.92}	\\
      ResNeXt29 	&	31.87	&	76.95	&	47.71	&	\textbf{88.57}	&	23.85	&	78.15	&	50.90	&	81.63	&	\textbf{35.04}	&	\textbf{81.87}	&	\textbf{56.92}	&	84.08	\\
      SENet18	&	29.18	&	78.66	&	42.16	&	90.17	&	25.63	&	\textbf{80.55}	&	\textbf{54.18}	&	91.71	&	\textbf{31.96}	&	79.90	&	51.69	&	\textbf{93.07}	\\
      ShuffleNetG2 	&	22.01	&	71.67	&	\textbf{48.93}	&	85.54	&	17.27	&	70.58	&	38.40	&	81.05	&	\textbf{27.72}	&	\textbf{74.56}	&	48.85	&	\textbf{91.14}	\\
      ShuffleNetV2 	&	24.89	&	72.95	&	49.26	&	87.44	&	20.50	&	71.25	&	46.44	&	86.18	&	\textbf{27.10}	&	\textbf{75.27}	&	\textbf{53.19}	&	\textbf{91.88}	\\
      VGG 	&	21.03	&	76.66	&	\textbf{43.41}	&	90.07	&	16.73	&	77.60	&	42.15	&	88.61	&	\textbf{23.16}	&	\textbf{80.75}	&	43.37	&	\textbf{93.06}	\\

			\cmidrule(l){1-13} 
		\end{tabular}
	}
	
	\label{tbl1}
\end{table*}

In the experiments, we also use four real datasets, 1- Tiny ImageNet \citep{Le2015TinyIV} includes $120000$ natural images in $200$ classes with each class having a training set of $500$ images a test set of $50$ images along with $50$ validation images. 2,3- CIFAR-10 and CIFAR-100 datasets \citep{krizhevsky2009learning}, both containing $60000$ images of size $32\times 32$ in $10$ and $100$ classes respectively. 4- SVHN \citep{netzer2011reading} which contains over $600000$ images of real-world images of digits $0-9$. These selected datasets are used for three main reasons. First, while they are complex datasets,  they have a reasonable number of images and features, which makes working with them with our available computational resources feasible. Second, they are well-known datasets with known and predictable results on a variety of ANN architectures.  Third,  they are compatible with the official experiments of AutoAugment paper  \citep{autoaugment}. 

The proposed method can be used with all architectures and all datasets with reasonable training size. For training, we have two completely separate steps,  1-finding policies, 2-applying those policies. For finding policies, we use Algorithm 1, and then we perform normal training with the new policies. The default learning rate for networks in \citep{kuangliu} is 0.1. We use the same learning rate throughout the experiments for finding policies and training networks. The number of epochs for all of the training scenarios was $200$.  To find the best policies, we need to create child networks. The child networks and training networks share the same infrastructures. To obtain the accuracies from child networks, we divided training data into two parts. The training part and the testing part. For Tiny Imagenet and SVHN, $5000$ images are used for testing. For  CIFAR-10 and CIFAR-100, $2500$ images are used for testing. The images which are not used for testing part are used for training part. All of the images are selected randomly with i.i.d. distribution. The $\alpha$ value for Pareto Distribution was always $2$. We also used $d=25$ and $\lambda=5$.

\subsection{Accuracy}

In this section, we test the accuracy of our proposed method using four different datasets, 1- Tiny ImageNet, 2- CIFAR-10, 3- CIFAR-100, and 4- SVHN.  The results are shown in Table \ref{tbl1}.  In this table, the "Manual" section stands for the images with common augmentation methods. These techniques include zero padding, cropping, random-flip, and cutout. To prevent probable errors, the same source code released in AutoAugment for augmentation techniques is also used for manual augmentation  \citep{autoaugment}. This helps to provide a fair environment for all methods. The only extra pre-processing step that we used is for resizing Tiny Imagenet from $64 \times 64$ to $32 \times 32$. This helped us to use the same infrastructure for all datasets without having adverse effects on the experiments. The values in the table are the average results from five trials.

The section "AutoAugment" stands for the AutoAugment method. For CIFAR-10, CIFAR-100, and SVHN, we use the same policies that are found from AutoAugment method. For Tiny ImageNet, we use the policies that are found for ImageNet dataset. Because Tiny ImageNet is a subset of ImageNet, it can test the generalization of the AutoAugment method. The section  "GAutoAugment" stands for Greedy AutoAugment, which is the proposed method. In our method, we suggest a  specific searching process for each scenario to find the best policies. This is possible because (as we will see in the next section), our search method is computationally much more efficient than AutoAugment method.

For Tiny ImageNet, as we can see, the policies from AutoAugment are not effective when they are applied to a smaller subset of the same dataset. All of the networks had worse results compared to the Manual augmentation. Overall, the AutoAugment reduced the accuracy with $52.73 \%$ compared to Manual augmentation. Comparatively, our method increased accuracies for nine out of eleven available networks. The highest increase of accuracy is for ResNet18, with $6.23\%$ higher accuracy. The least increase of accuracy is for VGG with $2.12\%$ higher accuracy. Overall, compared to the AutoAugment, the proposed method provided $78.34\%$ higher accuracy for eleven networks. Respectively, compared to the Manual augmentation, the proposed method provided $25.60\%$ higher accuracy for eleven networks.

For CIFAR-10, the transition is better for AutoAugment policies. From eleven networks, six networks had better results with at most $5.89\%$ better accuracy and at least $0.06\%$ better accuracy than the manual augmentation. On the other hand, the manual augmentation showed better results for five networks with at most $4.54\%$ better accuracy for MobileNet and at least $0.46\%$ better accuracy in DenseNet121. The overall increase of the accuracy was $0.86\%$ in favor of the AutoAugment. Comparatively, our method increased the accuracies for nine networks, with at most $6.27\%$ better accuracy for PreActResNet18 and at least $1.24\%$ better accuracy for SENet18. Overall, we could increase the accuracies, $25.02\%$ better than manual augmentation and $24.15\%$ better than AutoAugment.

The results for CIFAR-100 show that when policies from AutoAugment were applied to original images, the accuracies could improve for six networks. The best increase of the accuracy is for GoogLeNet with at most $7.49\%$ better accuracy, and the least accuracy is for DenseNet121 with at least $0.91$ better accuracy. The overall increase of the accuracy was $1.04\%$ for AutoAugment compared to the manual augmentation. When the policies from the proposed method are applied to the original network, we could improve the results for nine networks. The accuracy could be up to $11.54\%$ and down to $2.34\%$ better than the original images. The overall increase of the accuracy for the proposed method is $56.46\%$ compared to the manual augmentation. The overall increase of the accuracy for all of the networks is $55.422\%$ for the proposed method compared to the AutoAugment.

For SVHN, the AutoAugment provide higher accuracies for four networks when it is compared to the manual augmentation. The highest accuracy is for ResNet18 with $2.75\%$ better accuracy and the least higher accuracy is $0.70\%$ for PreActResNet18. For the six other networks, the manual augmentation was, on average $2.45\%$ better than AutoAugment. It was at least $0.21\%$ and at most $6.93\%$ better than AutoAugment. Overall, manual augmentation provided $11.00\%$ better cumulative results compared to the AutoAugentation. Comparatively, the proposed method provided better accuracies for ten networks compared to the manual augmentation. The highest and lowest accuracies were $7.11\%, 1.57\%$ better than manual augmentation. Overall, the cumulative accuracies are $37.16\%,$ and $48.16\%$ better than manual augmentation, and AutoAugment. The better accuracy comes from the fact that SVHN needed policies that are in further layers of the search space, and this could only be achieved with Greedy AutoAugment.

\subsection{Computational Analysis}

As described in Section 3, the search space is reduced from the exponential growth of base $(t_n \times p_n \times m_n)$ with the value of $\ell$ to the linear growth of $(t_n \times m_n)$ with the value of $k$. We already saw that this approach is effective in improving the accuracy of the networks. The important question is now how much computational resources we needed to perform the search. To allocate the resources, we are concerned about two aspects, 1-the search space, 2-real resources allocated for the searching process. We separate this two processess because the algorithms may not use all of the trials is searching spaces.

\begin{table}[h]
	\centering
	\caption{The result table for comparing the search space between our method and AutoAugment.}
	\label{tbl2}
	\resizebox{75mm}{!}{
		{
			\begin{tabular}{|@{}l|ll|l@{}|}
				\hline
				Layers&AutoAugment   &GAutoAugment &Comparison \\
				\hline
				\hline
				$\ell=1$
				&$2200^1$&$200$  &$11.0$\\
				$\ell=2$       
				&$2200^2$&$4200$ &$1152.3$\\
				$\ell=3$  
				&$2200^3$&$8200$&$1298536.5$\\
				$\ell=4$  
				&$2200^4$&$12200$&$1920131147.5$\\
				\hline				
			\end{tabular}
		}
	}
	
\end{table}

The comparison between the two spaces is shown in Table~\ref{tbl2}. This table has three columns. The first column shows the number of the layers, which is ranged from $\ell=1$ to $\ell=4$. The second column is for the search space for the AutoAugment algorithm. This shows the maximum number of possible selections that AutoAugment can choose. As we can see, the values for each row exponentially increases with the increase of the $\ell$ from a base of $2200$. The $2200$ is from calculating $(t_n \times p_n \times m_n)$. The third column represents the search space for Greedy AutoAugment. To calculate the search space, we use the (2). In this case, if each layer is fully explored, the number of possible trials is only $4000$.  The last column shows the comparison between column two and column three. The values show that the number of times the search space of the proposed method is smaller than the search space of AutoAugment.

For the real experiments, AutoAugment only searches for the 2-layer search space. This means that AutoAugment can have $2200^2$ possible trials. Because searching all of the space for $2200^2$ trials is impractical, the AutoAugment uses a sample of $15000$  child networks. Comparatively, our method, in each step, explores $200$ trials. The number of steps is determined by $k$. In these experiments, we used $k=5$, which gives us 1000 trials. This number is used to limit the search within our computational resources. Also, AutoAugment uses $120$ epochs to evaluate the accuracies of child networks. The number of epochs used for our child networks was only  $5$ epochs.   Since the exact infrastructure for different datasets is not known for AutoAugment, and child networks are interchangeable,  we consider the child networks to have the same efficiency. Therefore, overall we had $(15000 \times 120) \div (1000 \times 5) = 360$ less computational requirement.

\section{Conclusion}

A major problem in data augmentation is to ensure that the generated new samples cover the search space. In this paper, we proposed Greedy AutoAugment as a highly efficient method to find the best augmentation policies. We combined the searching process with a simple procedure to add augmentation policies to the training data. For the experiments on the classification accuracy, we used four real datasets and eleven networks. Our results show that the proposed method could reliably improve the accuracy of classification results. The higher accuracy could be achieved while using 360 times less computational resources.   

\bibliographystyle{model2-names}
\bibliography{bibi}

\end{document}